%% file: main.tex
\documentclass[runningheads]{llncs}

\usepackage{eccv}

\usepackage{eccvabbrv}

\usepackage{graphicx}
\usepackage{booktabs}

\usepackage{multirow}
\usepackage{pifont}
\usepackage{float} 
\usepackage{diagbox}

\usepackage{colortbl}
\usepackage{pgf}  

\usepackage{subcaption}
\usepackage{tabularx}

\usepackage{makecell}
\usepackage{color} 

\usepackage{wrapfig}

\usepackage[accsupp]{axessibility}

\usepackage{hyperref}

\usepackage{orcidlink}

\makeatletter
\renewcommand{\@fnsymbol}[1]{%
  \ensuremath{%
    \ifcase#1\or
      \star\or
      \dagger\or
      \ddagger\or
      \mathchar "278\or
      \mathchar "27B\or
      \|\or
      \star\star\or
      \dagger\dagger\or
      \ddagger\ddagger
    \else
      \@ctrerr
    \fi}}
\makeatother

\begin{document}

\title{Beyond Atomic Layouts: Compositional Design Understanding with Vision-Language Models}

\titlerunning{Beyond Atomic Layout}

\author{Yiyang Huang\inst{1,2}\thanks{This work was done during an internship at Adobe Research.} \and
Zhaowen Wang\inst{2}\thanks{Project lead.} \and
Simon Jenni\inst{2} \and
Jing Shi\inst{2} \and \\
Yitian Zhang\inst{1} \and
Yizhou Wang\inst{1} \and
Yun Fu\inst{1}}

\authorrunning{Y.~Huang et al.}

\institute{
Northeastern University, Boston, USA \and
Adobe Research, San Jose, USA\\
\url{https://hukcc.github.io/Beyond-Atomic-Layouts/}
}

\maketitle

\input{sec/0_abstract}
\input{sec/1_intro}
\input{sec/2_related_work}
\input{sec/3_method}

\input{sec/4_exp}

\input{sec/5_conclusion}

\bibliographystyle{splncs04}
\bibliography{main}

\end{document}

%% file: sec/0_abstract.tex
\begin{abstract}
Layout understanding, or the interpretation of element organization, is essential for document analysis, user interface (UI) creation, and graphic design. While recent vision-language models (VLMs) excel at interpreting atomic layouts composed of independent elements, they struggle with compositional layouts that require reasoning over visually entangled elements within hierarchical multi-layer structures. 
In this paper, we introduce a new task, compositional layout understanding, and present CoDeLayout, a VQA dataset of $\sim$20K real-world multi-layer layouts annotated with compositional element pairs and design intent. Through empirical analysis on CoDeLayout, we identify two key challenges for existing VLMs: semantic drift between textual metadata and visual content, and structural ambiguity in hierarchical inter-element relationships. 
To address these challenges, we propose MASON, a post-training paradigm that integrates multimodal alignment (MA) and structural perception (SP). MA enhances element interpretation by grounding metadata-defined elements to their visual counterparts, mitigating semantic drift, while SP models layer-aware inter-element spatial relationships to improve hierarchical understanding and reduce structural ambiguity. 
Experiments reveal substantial gaps in existing VLMs: even the strongest baseline, GPT-o3, achieves only 79.68\% accuracy, whereas Qwen2.5-VL 7B with MASON reaches 91.66\%. Notably, MASON surpasses full-data Direct Finetune using only 30\% of the training data and scales better with additional data. 
\keywords{Compositional Layout \and Graphic Design \and Post-training}
\end{abstract}

%% file: sec/1_intro.tex
\input{figs/teaser}
\section{Introduction}
\label{sec:intro}

Layout understanding, the ability to perceive structural organization and relationships among elements such as text, images, and graphics, is a fundamental capability for intelligent systems in document analysis \cite{layoutlmv3}, user interface (UI) creation \cite{cogagent}, and graphic design \cite{AesthetiQ}.

Recent VLMs have shown strong performance in layout interpretation. In document understanding, models such as the LayoutLM series~\cite{layoutlm, layoutlmv2, layoutlmv3} integrate semantic and layout representations through 2D position embeddings, enabling fine-grained reasoning over individual document elements such as headers, tables, and forms. In user interface (UI) understanding, datasets such as ScreenQA~\cite{ScreenQA} enable models like CogAgent~\cite{cogagent} to learn navigation in graphical user interfaces (GUIs) through interaction-related visual question answering (VQA) at the level of individual interface elements. In graphic design, recent approaches address tasks such as extracting quantitative information from infographics~\cite{InfographicVQA} and analyzing rhetorical intent in advertisements~\cite{malakouti2025wacv} by modeling grouping relationships among individual layout elements. However, existing research is primarily developed for atomic layouts composed of independent elements, leaving compositional structures underexplored.

In contrast to atomic layouts, which typically involve clearly separated elements with explicit boundaries and rely on 2D positional encodings to model reading order, compositional designs involve visual entanglement among multiple elements within hierarchical structures and require layer-aware spatial reasoning to capture inter-element relationships. Figure~\ref{fig:teaser} (a-d) illustrates representative composition types, including \textit{clipping} for inter-layer insertion, \textit{blending} image crops into stylized shapes, \textit{overlaying} an element with its outline, and \textit{morphing} text for visual effects. Furthermore, as shown in Figure~\ref{fig:teaser_task}, understanding compositional layouts is essential for enabling design assistants to perform editing operations correctly. This, in turn, requires the underlying VLMs to interpret elements under visual and semantic ambiguity arising from multi-element entanglement and to reason over layer-aware inter-element relationships. Accordingly, we introduce compositional layout understanding as a new task for VLMs and present CoDeLayout, a VQA dataset designed to support it. Comprising approximately 20K real multi-layer design layouts, CoDeLayout provides high-fidelity rendered images paired with element-level metadata and QA-style annotations that capture compositional element pairs and their associated design intents.

Empirical analysis on CoDeLayout reveals that existing VLMs struggle with compositional layouts due to two key challenges: \textit{semantic drift} and \textit{structural ambiguity}. Semantic drift emerges from the visual entanglement of elements, where VLMs fail to align textual metadata with visual content, leading to incorrect interpretation of element semantics and design intent. Structural ambiguity stems from the hierarchical organization of compositional elements, where VLMs struggle to interpret layer-aware spatial structures and consequently fail to capture inter-element relationships.

To address these challenges, we propose \textbf{MASON}, a post-training paradigm integrating \textbf{M}ultimodal \textbf{A}lignment (MA) and \textbf{S}tructural percepti\textbf{ON} (SP). MA mitigates semantic drift by aligning metadata-defined elements to their visual counterparts, enabling coherent interpretation of design attributes and visual semantics. In parallel, SP addresses structural ambiguity by extracting layer-aware spatial relationships among elements that encode their positional and hierarchical dependencies.

Experiments reveal substantial gaps in existing VLMs on compositional layout understanding: even the strongest baseline, GPT-o3 with image-based reasoning capability, achieves only 79.68\% accuracy. In contrast, Qwen2.5-VL 7B with MASON achieves 91.66\% accuracy and consistently improves performance across diverse compositional design categories, including overlaying, clipping, blending, and morphing. Furthermore, MASON demonstrates strong data efficiency, surpassing full-data Direct Finetune while using only 30\% of the training data, and shows better scalability as training data increases.

Our main contributions are summarized as follows:
\begin{itemize}
\item We introduce compositional layout understanding as a new task and construct CoDeLayout, the first VQA dataset for this setting, comprising $\sim$20K real multi-layer layouts with explicit annotations of compositional element pairs and their design intent.
\item Through empirical analysis on CoDeLayout, we identify two core challenges, namely \textit{semantic drift} and \textit{structural ambiguity}, explaining why existing VLMs struggle with compositional layouts.
\item We propose MASON, a post-training paradigm that mitigates semantic drift via multimodal alignment and addresses structural ambiguity through layer-aware structural perception.
\item Experiments reveal substantial performance gaps in existing VLMs and validate the effectiveness of MASON. Notably, MASON surpasses full-data Direct Finetune using only 30\% of the training data and scales more effectively with increasing data.
\end{itemize}

%% file: figs/teaser.tex
\begin{figure}[t]
    \centering

    \begin{minipage}{0.96\linewidth}
        \centering

        \begin{subfigure}{0.48\linewidth}
            \centering
            \includegraphics[width=0.925\linewidth]{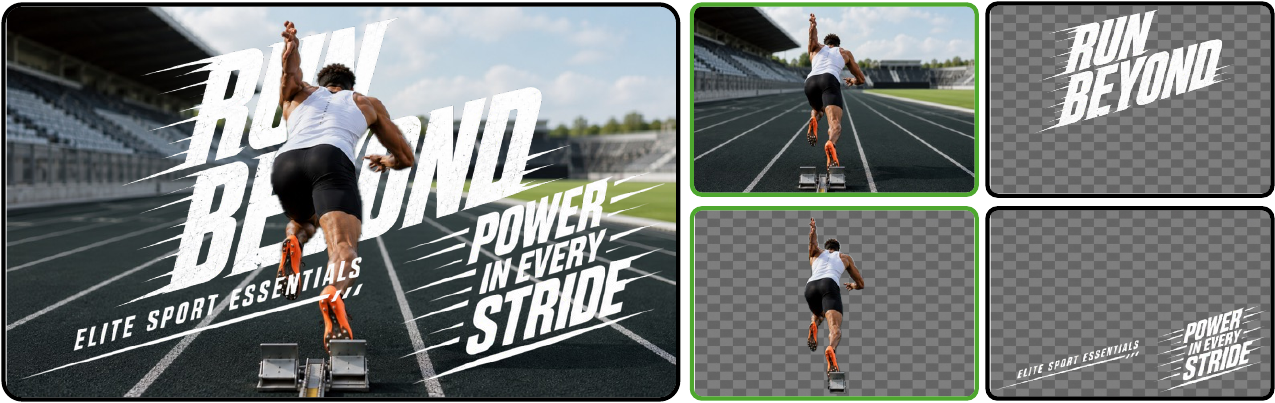}

            \caption{Clipping: cutout layer for insertion}
            \label{fig:teaser_clipping}
        \end{subfigure}\hfill
        \begin{subfigure}{0.48\linewidth}
            \centering
            \includegraphics[width=0.925\linewidth]{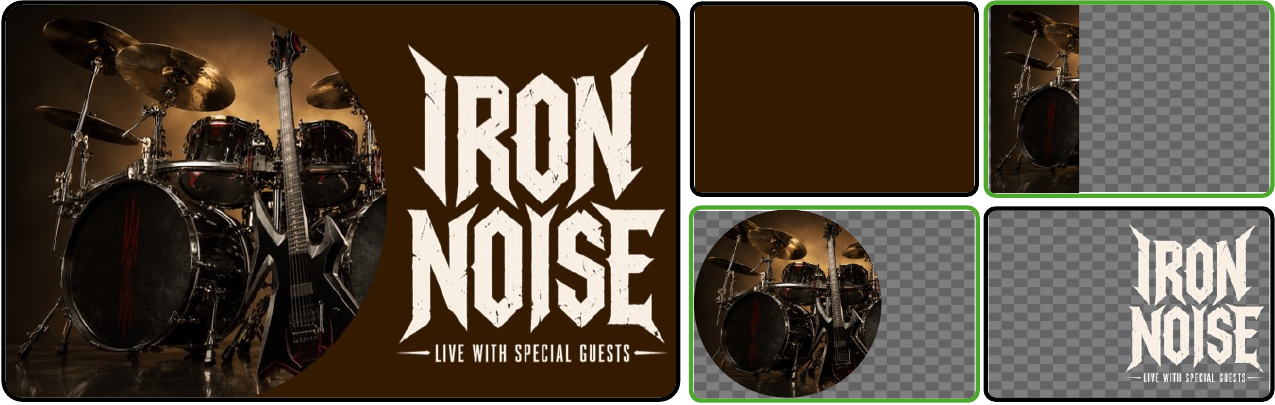}
            \caption{Blending: merged image crops}
            \label{fig:teaser_blending}
        \end{subfigure}

        \begin{subfigure}{0.48\linewidth}
            \centering
            \includegraphics[width=0.925\linewidth]{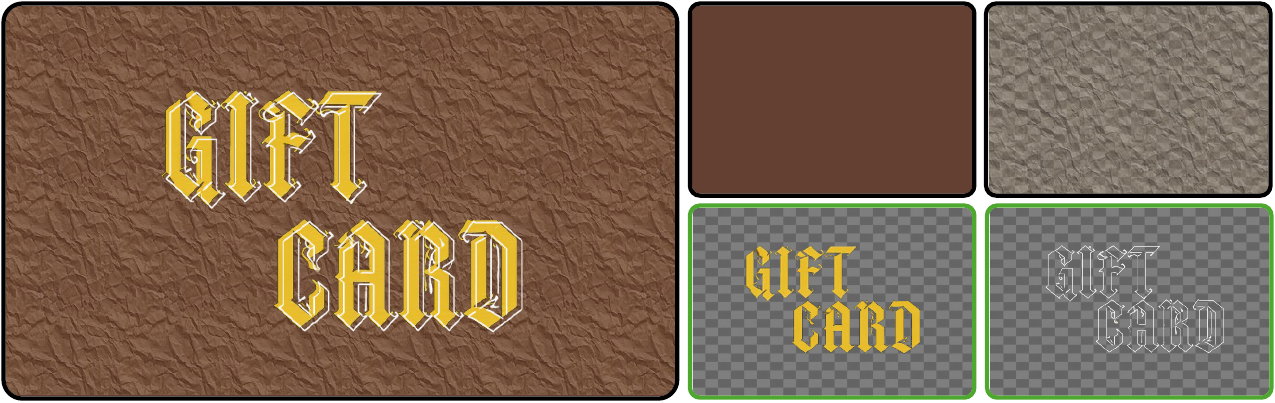}

            \caption{Overlaying: element + outline/shadow}
            \label{fig:teaser_overlaying}
        \end{subfigure}\hfill
        \begin{subfigure}{0.48\linewidth}
            \centering
            \includegraphics[width=0.925\linewidth]{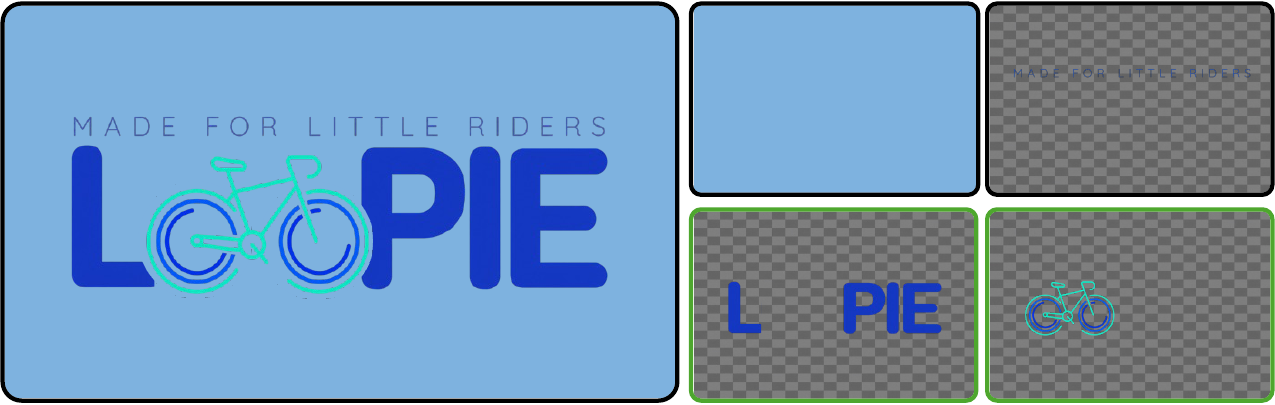}

            \caption{Morphing: text replaced by graphics}
            \label{fig:teaser_morphing}
        \end{subfigure}

        \begin{subfigure}{\linewidth}
            \centering
            \includegraphics[width=0.975\linewidth]{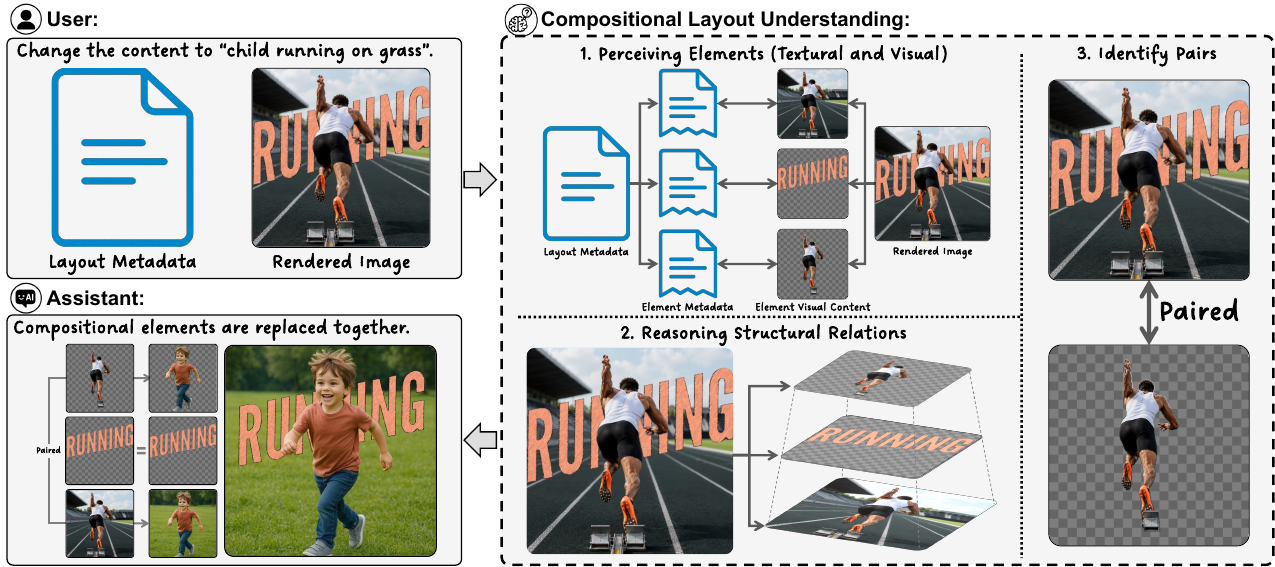}
            \caption{Compositional layout understanding for design assistants}
            \label{fig:teaser_task}
        \end{subfigure}
    \end{minipage} 

    \caption{
    \textbf{Compositional layouts in graphic design.}
    \textbf{(\subref{fig:teaser_clipping}--\subref{fig:teaser_morphing})} show representative compositional layouts, with green boxes marking key elements.
    \textbf{(\subref{fig:teaser_clipping})} Clipping extracts a region from a base image as a separate layer, allowing insertion of new elements between the base and the cutout;
    \textbf{(\subref{fig:teaser_blending})} Blending integrates different crops of the same image into stylized shapes for aesthetic coherence;
    \textbf{(\subref{fig:teaser_overlaying})} Overlaying combines an element with its outline or shadow to emphasize visual hierarchy; and
    \textbf{(\subref{fig:teaser_morphing})} Morphing replaces text with graphics to create expressive visual effects.
    \textbf{(\subref{fig:teaser_task})} Compositional layout understanding is critical for design assistants, as handling such layouts often requires coordinated edits across interrelated elements to faithfully implement user intent.
    }
    \label{fig:teaser}
\end{figure}

%% file: sec/2_related_work.tex
\section{Related Work}
We review related work in three areas: vision-language models as the foundation for layout reasoning, layout generation methods for design synthesis, and layout understanding approaches primarily focused on atomic layouts.

\subsection{Vision Language Models}

Recent advances in vision-language models (VLMs) ~\cite{mllm-survey, Vqtoken, ThinkJEPA, Indra, representation-survey, dcode} have substantially improved multimodal understanding by integrating large language models (LLMs)~\cite{Qwen,Learners,vicuna_report,PaLM,ChatGPT,LLaMA} with visual encoders such as CLIP~\cite{clip}. Flamingo~\cite{Flamingo} introduced interleaved vision-language modeling for open-ended reasoning, while BLIP-2~\cite{blip2} bridged frozen vision and language backbones through a Q-Former. Instruction-tuned models such as LLaVA~\cite{llava,llavanext} and Qwen-VL~\cite{Qwen-VL,Qwen2-VL} further improved visual grounding and cross-modal reasoning. Building on these advances, spatially aware VLMs such as Qwen2.5-VL~\cite{Qwen25-VL}, Qwen3-VL~\cite{Qwen3-VL}, InternVL-3.5~\cite{internvl35}, and LLaVA-OneVision~\cite{llava-onevision} further advance dense localization and object-level grounding. Collectively, these advances provide a strong foundation for layout perception and reasoning.

\subsection{Layout Generation}
Layout generation aims to synthesize new designs from learned structural patterns and has progressed across documents, user interfaces, and graphic design. For documents, methods such as LayoutVAE~\cite{LayoutVAE}, LayoutGAN++~\cite{layoutgan++}, LayoutDM~\cite{layoutdm}, and Text2Poster~\cite{Text2Poster} generate layouts that follow logical and visual hierarchies. For user interfaces, Pix2Code~\cite{pix2code} and Design2Code~\cite{Design2Code} synthesize coherent layouts from textual descriptions or sketches. In graphic design, datasets including Magazine~\cite{Magazine}, GenPoster-100K~\cite{GenPoster-100K}, DesignerIntention~\cite{cole}, Crello~\cite{Crello}, and MLTD~\cite{art} support multi-layer layout modeling, enabling models such as LayoutGAN~\cite{layoutgan}, LayoutTransformer~\cite{layouttransformer}, COLE~\cite{cole}, and ART~\cite{art} to learn geometric constraints and generate layouts under multimodal guidance.

\subsection{Layout Understanding}
Layout understanding aims to perceive and reason over the spatial and semantic organization of elements and is central to document analysis, UI creation, and graphic design. In documents, datasets such as PubLayNet~\cite{PubLayNet} and DocLayNet~\cite{DocLayNet} support models like LayoutLM~\cite{layoutlm,layoutlmv2,layoutlmv3} and DiT~\cite{dit}, which encode semantic and layout features with 2D positional embeddings. For UIs, RICO~\cite{Rico} and ScreenQA~\cite{ScreenQA} enable navigation in graphical user interfaces (GUIs) and interaction-related visual question answering (VQA) with models such as CogAgent~\cite{cogagent}. In graphic design, prior work addresses tasks including infographic question answering~\cite{InfographicVQA,ChartQA} and rhetorical analysis in advertisements~\cite{malakouti2025wacv} by modeling grouping relationships among individual elements.

Although effective for atomic layouts, these datasets and methods assume independent elements and rely on 2D positional encodings, limiting their ability to model hierarchical and layer-aware interactions among visually entangled elements. Such interactions are fundamental to compositional layout understanding and motivate the need for a dedicated dataset and a tailored modeling paradigm.

%% file: sec/3_method.tex
\section{Compositional Layout Understanding: Task \& Dataset}

\input{figs/dataformat}
\subsection{Task Formulation}
Compositional layout understanding aims to identify interacting elements within hierarchical layouts and interpret their design intent. As illustrated in Fig.~\ref{fig:dataset_examples}, each layout instance is represented as $\mathcal{L} = (\mathcal{I}, \mathcal{M})$, where $\mathcal{I}$ is the rendered design image and $\mathcal{M} = \{\mathcal{M}_{e_1}, \mathcal{M}_{e_2}, \dots, \mathcal{M}_{e_n} \}$ denotes the complete metadata of the design. Each $\mathcal{M}_{e_i}$ contains attributes of element $e_i$, including its type, spatial position (placement), and stacking order (Zindex). Given a question $Q(e^q)$ referring to a query element $e^q$, the model is required to identify the corresponding compositional element $e^{comp}$ and generate a textual answer $A(e^{comp})$ explaining their design intent:
\begin{equation}
A(e^{comp}) = \text{VLM}(\mathcal{L}, Q(e^q))
\end{equation}

\subsection{Compositional Design Layout Dataset}
\label{sec:dataset}

To support compositional layout understanding in graphic design, we construct the Compositional Design Layout Dataset (CoDeLayout), a high-quality, multi-layer design dataset centered on element-level compositional relationships. Detailed data collection procedures are provided in the supplementary materials.

\subsubsection{Data Format \& Statistics}

CoDeLayout contains approximately 20K design instances spanning diverse resolutions and coherent multi-layer structures across four representative compositional types: Overlaying, Clipping, Blending, and Morphing. As illustrated in Fig.~\ref{fig:dataset_examples}, each instance includes: (1) a high-fidelity rendered design image; (2) a structured JSON file with element-level metadata (\texttt{id}, \texttt{type}, \texttt{opacity}, \texttt{zIndex}, \texttt{position}, etc.); and (3) QA-style annotations $(Q(e^q), A(e^{comp}))$ specifying compositional element pairs $(e^q, e^{comp})$ and their associated design intents.

CoDeLayout follows the natural distribution of real-world graphic designs without resampling for category balancing. It contains 20,009 training samples and 387 test samples. In the training set, Blending dominates (56.78\%), followed by Overlaying (22.83\%), Clipping (17.76\%), and Morphing (2.63\%). To ensure reliable evaluation across all composition types, the test split slightly increases the proportion of rare categories, resulting in the following distribution: Blending (40.06\%), Overlaying (29.95\%), Clipping (17.75\%), and Morphing (12.24\%). All test samples are manually verified to ensure annotation correctness and prevent data leakage. In terms of resolution, CoDeLayout spans from small banners (300×60) to ultra-high-resolution prints (7200×14400), covering common social media formats (e.g., 1080×1080, 1080×1920) and standard print sizes (e.g., A4 at 2480×3507). The designs contain an average of 17.4 layers, with 99.8\% containing fewer than 45 layers.

\input{tables/dataset_compare}
\subsubsection{Comparison with Existing Layout Datasets}
Table~\ref{tab:dataset_comparison} compares representative layout datasets across understanding and generation domains. Document and UI datasets, such as PubLayNet~\cite{PubLayNet}, DocLayNet~\cite{DocLayNet}, ScreenQA~\cite{ScreenQA}, and RICO~\cite{Rico}, primarily target atomic layouts with flat or tree-structured representations, supporting structural parsing and function-level reasoning. Graphic design datasets, including Crello~\cite{Crello}, GenPoster-100K~\cite{GenPoster-100K}, and MLTD~\cite{art}, provide multi-layer layouts for generation but do not explicitly model inter-element compositional relationships.

In contrast, CoDeLayout is the first dataset dedicated to compositional layout understanding. It integrates high-fidelity rendered designs, element-level metadata, and explicitly formulated QA-style annotations that capture complex compositional element pairs and their associated design intent.

\subsection{Challenges in Compositional Layout Understanding}
\label{sec:challenge_analysis}
Compositional layouts introduce structural complexity beyond atomic layouts, requiring accurate element interpretation under visual entanglement (\textit{semantic drift}) and layer-aware modeling of hierarchical relationships among elements (\textit{structural ambiguity}). To examine these challenges, we analyze a strong VLM, GPT-4o, on CoDeLayout.

\input{figs/problem_def}
\subsubsection{Semantic Drift}
\label{sec:semantic_drift}
Semantic drift arises from the visual entanglement of elements, as VLMs fail to align textual metadata with visual content, resulting in misinterpretation of element semantics and degraded compositional understanding.

To empirically examine this issue, we evaluate GPT-4o’s ability to ground metadata-defined elements in the rendered layout image. For each layout, element crops are fed to GPT-4o with a fixed descriptive prompt (``What is element A?'') to generate reference descriptions. During evaluation, the model receives the rendered layout image together with its metadata and is asked to describe each element. Performance is measured by the average BLEU score between generated and reference descriptions across elements within each layout. 

As shown in Figure~\ref{fig:problem-vis-sub1}, samples with higher grounding scores consistently achieve better compositional identification accuracy. This result indicates that grounding capability directly impacts compositional layout understanding, highlighting multimodal alignment as a critical factor.

\subsubsection{Structural Ambiguity}
\label{sec:structural_ambiguity}
Structural ambiguity arises from the hierarchical organization of compositional elements, where VLMs struggle to interpret layer-aware spatial structures, limiting their ability to model inter-element relationships.

To empirically examine this issue, we evaluate GPT-4o’s ability to perceive layer-aware spatial relationships through a spatial relation perception task. For each layout, 30 element pairs are randomly sampled to construct yes/no questions about structural relationships (e.g., ``Is element A above element B?''). During evaluation, the model receives the rendered layout image together with its metadata and answers these questions. Performance is measured by the average accuracy across the sampled element pairs within each layout.

As shown in Figure~\ref{fig:problem-vis-sub2}, higher spatial relation perception accuracy consistently corresponds to better compositional identification accuracy. This indicates that structural perception directly affects compositional layout understanding.

\input{figs/pipeline}
\section{MASON: A Baseline Post-training Paradigm}

Our analysis in Section~\ref{sec:challenge_analysis} shows that semantic drift and structural ambiguity hinder accurate element interpretation and inter-element relationship modeling, limiting VLMs’ compositional layout understanding. To address these challenges, we propose MASON as a baseline paradigm for compositional layout understanding that integrates multimodal alignment (MA) and structural perception (SP) into VLM post-training, as illustrated in Figure~\ref{fig:pipeline}.

\subsection{Multimodal Alignment (MA)}
To address semantic drift, MA introduces an element grounding objective during VLM post-training, encouraging alignment between textual metadata and corresponding visual content under visual entanglement.

The grounding objective is constructed from 1K layouts sampled from the training split of CoDeLayout. Each instance follows the layout formulation $\mathcal{L} = (\mathcal{I}, \mathcal{M})$ and is augmented with grounding QA supervision $(Q^g, A^g)$.

Specifically, for each layout, 20\% of elements are randomly sampled. For each selected element $e_i$, an off-the-shelf VLM $\mathcal{G}$ generates a grounding QA pair. Given a layout $\mathcal{L}$, together with the element crop $\mathcal{I}_{e_i}$, its metadata $\mathcal{M}_{e_i}$, and a grounding question $Q^g(e_i)$ (``What is the element with ID \texttt{\{element\_id\}} and where is it located in the complete design?''), $\mathcal{G}$ produces an answer $A^g(e_i)$ describing its semantic identity and spatial placement:
\begin{equation}
A^g(e_i) = \mathcal{G}\big(\mathcal{P}_{\text{ground}}(\mathcal{L}, \mathcal{I}_{e_i}, \mathcal{M}_{e_i}, Q^g(e_i))\big),
\end{equation}
where $\mathcal{P}_{\text{ground}}$ denotes the grounding prompt template (Table~\ref{tab:prompt_element_grounding}).

\input{tables/grounding_prompt}

\subsection{Structural Perception (SP)}
To address structural ambiguity, SP incorporates layer-aware spatial relationships into VLM post-training, enabling accurate modeling of hierarchical inter-element relationships.

Given a layout $\mathcal{L}$ with metadata $\mathcal{M} = \{\mathcal{M}_{e_1}, \dots, \mathcal{M}_{e_n}\}$, where each element $e_i$ is annotated with attributes such as bounding box, stacking order, and type, SP derives spatial relationships between the query element $e^q$, referenced in the compositional layout understanding question $Q(e^q)$, and every other element $e_i$. These relationships are encoded in an additional metadata field \texttt{spatialRelationship}, which includes six properties: overlap status and ratio, containment type, center distance, directional relation, and relative stacking order (Table~\ref{tab:spatial_schema_json}). The enriched metadata for each element is defined as:
\begin{equation}
\mathcal{M}_{e_i}^{\prime} = \mathcal{M}_{e_i} \cup \texttt{spatialRelationship}(e_i, e^q).
\end{equation}

Embedding these spatial attributes into the metadata provides explicit geometric context, facilitating interpretation of layout hierarchy and inter-element structure.

\input{tables/spatial_info_json}
\subsection{Integration into VLM Post-training}

During post-training, multimodal alignment and structural perception are incorporated through metadata augmentation and joint supervision from compositional and grounding QA tasks.

Specifically, the original metadata $\mathcal{M}$ is augmented with layer-aware spatial relationships, resulting in $\mathcal{M}^{\prime} = \{\mathcal{M}_{e_1}^{\prime}, \dots, \mathcal{M}_{e_n}^{\prime} \}$, where each $\mathcal{M}_{e_i}^{\prime}$ includes the \texttt{spatialRelationship} relative to the query element $e^q$. The VLM is optimized on a mixture of compositional and grounding QA samples:
\begin{equation}
\hat{A} = \text{VLM}\big((\mathcal{I}, \mathcal{M}^{\prime}),\; Q\big),
\end{equation}
where $Q$ denotes either a compositional question $Q(e^q)$ or a grounding question $Q^g(e_i)$ depending on the training instance. The output is optimized using cross-entropy loss against the corresponding ground-truth answer.

At inference time, layer-aware metadata augmentation is retained, while evaluation focuses solely on the compositional layout understanding task.

%% file: figs/dataformat.tex
\begin{figure}[t]
    \centering
    \includegraphics[width=1.0\linewidth]{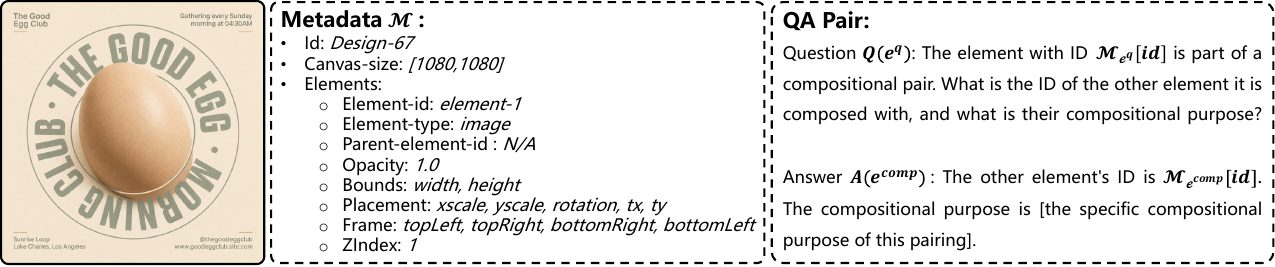}
    \caption{
    \textbf{Data format of a compositional design.} Each instance includes a rendered layout image, element-level metadata, and QA-style annotations specifying compositional element pairs and their associated design intents.
    }

    \label{fig:dataset_examples}
\end{figure}

%% file: tables/dataset_compare.tex
\begin{table}[t]
\centering

\caption{Comparison of representative layout datasets. In contrast to existing datasets that primarily focus on atomic layouts or generative objectives, CoDeLayout provides multi-layer layouts with explicit annotations of inter-element compositional relationships and the design intent.}

\scriptsize

\begin{tabular*}{\linewidth}{@{\extracolsep{\fill}}lccccc@{}}
\toprule
\textbf{Dataset} & \textbf{Annotation} & \textbf{Structure} & \textbf{Domain} & \textbf{\# Samples} & \textbf{\# Layers} \\
\midrule
\multicolumn{6}{l}{\textbf{Layout Generation}} \\
\midrule
RICO~\cite{Rico} & N/A & Flat & UI & $\sim$72K & N/A \\
Magazine~\cite{Magazine} & N/A & Layering & Graphic & $\sim$3.9K & N/A \\
GenPoster-100K~\cite{GenPoster-100K} & N/A & Layering & Graphic & $\sim$100K & 2--20 \\
DesignerIntention~\cite{cole} & N/A & Layering & Graphic & $\sim$100K & 3--10 \\
Crello~\cite{Crello} & N/A & Layering & Graphic & $\sim$20K & 2--50 \\
MLTD~\cite{art} & N/A & Layering & Graphic & $\sim$1M & 2--50 \\
\midrule
\multicolumn{6}{l}{\textbf{Layout Understanding}} \\
\midrule
PubLayNet~\cite{PubLayNet} & Atomic & Flat & Doc & $\sim$360K & N/A \\
DocLayNet~\cite{DocLayNet} & Atomic & Flat & Doc & $\sim$80K & N/A \\
ScreenQA~\cite{ScreenQA} & Atomic & Flat & UI & $\sim$86K & N/A \\
\textbf{CoDeLayout (Ours)} & Compositional & Layering & Graphic & $\sim$20K & 3--50 \\
\bottomrule
\end{tabular*}

\label{tab:dataset_comparison}
\end{table}

%% file: figs/problem_def.tex
\begin{figure}[t]
    \centering

    \begin{subfigure}[t]{0.49\linewidth}
        \centering
        \includegraphics[width=\linewidth]{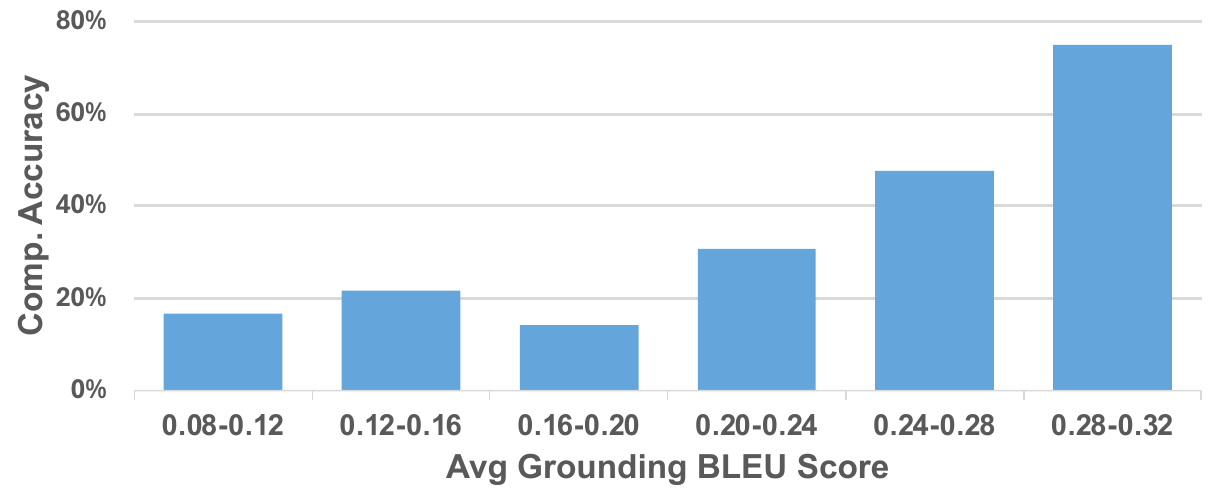}
        \caption{
        \textbf{Semantic Drift.}
        Grounding performance (x-axis: average BLEU score per layout) vs. compositional element identification accuracy (y-axis). 
        }
        \label{fig:problem-vis-sub1}
    \end{subfigure}%
    \hfill%
    \begin{subfigure}[t]{0.49\linewidth}
        \centering
        \includegraphics[width=\linewidth]{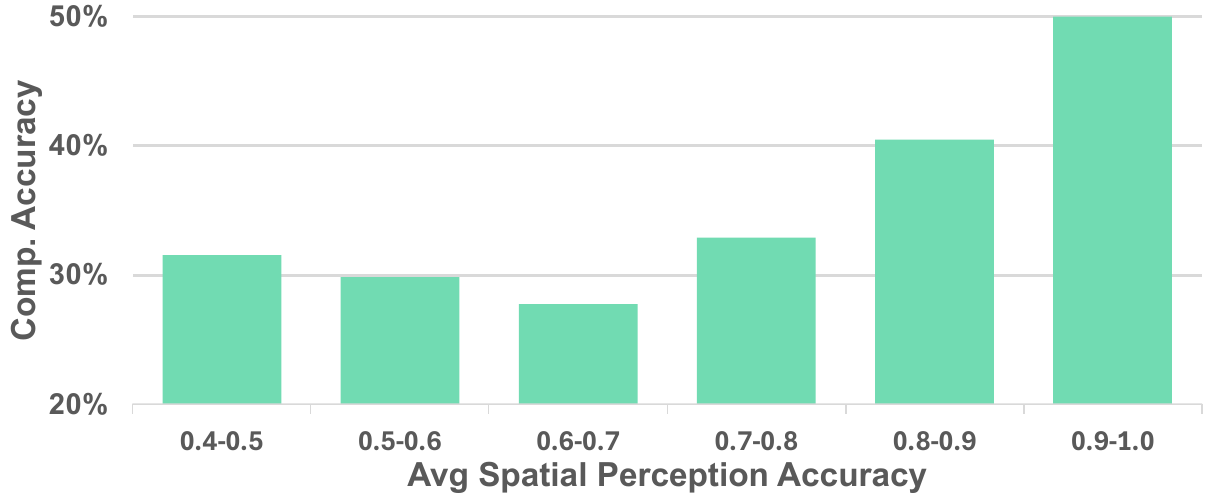}
        \caption{
        \textbf{Structural Ambiguity.}
        Spatial relation perception accuracy (x-axis) vs. compositional element identification accuracy (y-axis).
        }
        \label{fig:problem-vis-sub2}
    \end{subfigure}

    \caption{
    Impact of semantic drift (a) and structural ambiguity (b) on compositional layout understanding. 
    Higher grounding performance and spatial relation perception accuracy consistently correspond to better compositional element identification accuracy, indicating that multimodal alignment and structural perception are essential for compositional layout understanding.
    }
    \label{fig:problem-vis}
\end{figure}

%% file: figs/pipeline.tex
\begin{figure}[t]
    \centering
    \includegraphics[width=1.0\linewidth]{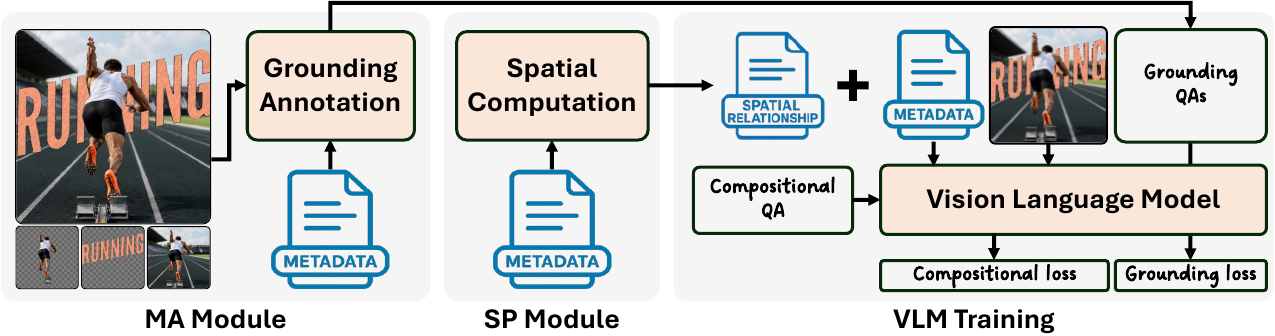}
    \caption{\textbf{Overview of MASON.} MASON mitigates semantic drift and structural ambiguity through multimodal alignment (MA) and structural perception (SP). MA introduces an element grounding objective to align textual metadata with visual content under visual entanglement. SP augments metadata with layer-aware spatial relationships, enabling accurate modeling of hierarchical inter-element structures.}
    \label{fig:pipeline}
\end{figure}

%% file: tables/grounding_prompt.tex
\begin{table}[t]
\centering

\caption{Prompt template $\mathcal{P}_{\text{ground}}$ for generating element grounding QA pairs in multimodal alignment.}

{\fontsize{7pt}{9pt}\selectfont
\begin{tabular}{p{12cm}}
\toprule
\textbf{Element Grounding Prompt $\mathcal{P}_{\text{ground}}$} \\ 
\midrule

\textbf{Inputs:} \\
- Full design image: $\mathcal{I}$ \\
- Layout metadata: $\mathcal{M} = \{\mathcal{M}_{e_1}, \mathcal{M}_{e_2}, \dots, \mathcal{M}_{e_n}\}$ \\
- Element ID: \texttt{\{element\_id\}} \\
- Image crop and metadata of the element: $(\mathcal{I}_{e}, \mathcal{M}_e)$ \\
\\

\textbf{Grounding Question $Q^g(e_i)$:} \\
``What is the element with ID \texttt{\{element\_id\}} and where is it located in the complete design?'' \\
\\

\textbf{Example Answer:} \\
``The element with ID \texttt{\{element\_id\}} is a [description]. It is located in the [position] of the design. [Optional: Additional identifying details].'' \\
\\

Please generate the answer based on both the element's metadata and visual content, as well as the global layout context. \\

\bottomrule
\end{tabular}
\label{tab:prompt_element_grounding}}
\end{table}

%% file: tables/spatial_info_json.tex
\begin{table}[t]
\centering

{\fontsize{7pt}{9pt}\selectfont
\caption{Schema of layer-aware inter-element spatial relationships used for structural perception.}
\begin{tabular}{p{12cm}}
\toprule
\textbf{Spatial Relationship Schema} \\ 
\midrule

\texttt{spatialRelationship: \{} \\
\hspace{1em}\texttt{hasOverlap: <True $\mid$ False>,} 
\hspace{1em}\texttt{overlapPercentage: <float>,} \\
\hspace{1em}\texttt{containment: <contain $\mid$ contained $\mid$ none>,} 
\hspace{1em}\texttt{centerDistance: <float>,} \\
\hspace{1em}\texttt{relativePosition: <left $\mid$ right $\mid$ above $\mid$ below>,} 
\hspace{1em}\texttt{layerOrder: <above\_given $\mid$ below\_given>} \\
\texttt{\}} \\

\bottomrule
\end{tabular}
\label{tab:spatial_schema_json}}
\end{table}

%% file: sec/4_exp.tex
\section{Experiments}

\subsection{Baselines and Metrics}

\subsubsection{Baselines.}
We evaluate heuristic methods, open-source models, and proprietary large-scale models on CoDeLayout to assess compositional layout understanding in current VLMs under zero-shot and default inference settings. Heuristic baselines include Max-Overlap, which selects the element with the highest IoU with the query $e^q$, and Nearest-Neighbor, which selects the element with the nearest centroid. Open-source models (7B scale) include Qwen2.5-VL~\cite{Qwen25-VL}, Qwen3-VL~\cite{Qwen3-VL}, InternVL-3.5~\cite{internvl35}, and LLaVA-OneVision~\cite{LLaVAOneVision}. Proprietary models include GPT-4o~\cite{GPT4o}, GPT-5~\cite{openai_gpt5_2025}, GPT-o3~\cite{openai_gpto3_2025}, Gemini-2.5-flash~\cite{Gemini25}, and Gemini-2.5-pro~\cite{Gemini25}. Models with explicit reasoning capabilities (GPT-5, GPT-o3, and Gemini-2.5-pro) are evaluated under their default medium reasoning configuration.

For post-training comparison, we further evaluate two Qwen2.5-VL~\cite{Qwen25-VL} 7B variants: Direct Finetune, finetuned solely on compositional QA data, and MASON, optimized with multimodal alignment and structural perception.

\subsubsection{Metrics.}
We report Accuracy, GPT-Score (0–10), BLEU, and ROUGE. Accuracy evaluates correct identification of the compositional element $e^{\text{comp}}$ given the query $e^q$, reflecting compositional element selection. For generated explanations, BLEU~\cite{bluescore} and ROUGE~\cite{rougescore} measure lexical overlap with reference answers, while GPT-Score~\cite{GPTscore} assesses semantic alignment with the annotated design intent. Detailed scoring prompts are provided in the supplementary materials.

\subsection{Implementation Details}

All models are evaluated under consistent prompting, decoding, and input configurations. Open-source VLMs (7B scale) are tested using their official instruction-following interfaces in zero-shot mode, while proprietary models are accessed via API under default inference settings. Input images are resized to $336\times336$ and paired with structured JSON metadata. All models use greedy decoding with a maximum output length of 256 tokens to ensure deterministic and parsable responses.

Our post-training variants, Direct Finetune and MASON, are implemented on Qwen2.5-VL (7B)~\cite{Qwen25-VL} and trained on CoDeLayout. Training uses the AdamW optimizer~\cite{AdamW} with a learning rate of $3\times10^{-5}$. LoRA adapters~\cite{LoRA} ($r=4$, $\alpha=8$) are applied to the attention projection layers (\texttt{q}/\texttt{v}). The context length is set to 4096 tokens, and post-training is conducted for 3 epochs with a global batch size of 64 on 8 A100 GPUs. For multimodal alignment, GPT-4o is used to generate grounding QA supervision. 
In experiments with type-balanced subsampling, abundant types (e.g., Overlaying, Blending, and Clipping) are sampled less frequently than scarce types (e.g., Morphing).

\input{tables/main-results}
\subsection{Comparison on CoDeLayout}

As shown in Table~\ref{tab:main_results}, heuristic baselines perform poorly across categories, indicating that simple geometric priors (e.g., overlap or centroid proximity) are insufficient for compositional element identification. 
Open- and closed-source VLMs also exhibit clear performance gaps under zero-shot settings. Even the strongest baseline, GPT-o3, achieves 79.68\% weighted accuracy and 76.28\% average accuracy, underscoring the difficulty of modeling visually entangled and layer-aware structures.

MASON consistently outperforms all baselines. Compared with GPT-o3, MASON (Full data) improves weighted accuracy from 79.68\% to 91.66\% and average accuracy from 76.28\% to 88.07\%, with especially large gains on structurally complex categories such as Clipping (90.91\% vs. 63.64\%), where layer-aware spatial modeling under visual entanglement is essential. 
Notably, even MASON achieves only 70.21\% accuracy on the challenging Morphing category, which is characterized by strong visual-semantic disruption and limited training data ($\sim$3\%), suggesting that CoDeLayout remains challenging.

MASON also shows strong data efficiency by achieving 89.32\% weighted and 86.12\% average accuracy, surpassing Direct Finetune trained on the full dataset (88.80\% and 84.30\%) while using fewer total QA pairs ($\sim$8K vs. 20K) under the same post-training protocol. Specifically, MASON uses only $\sim$6K compositional pairs (30\% of the training set, obtained via type-balanced subsampling) and $\sim$2K grounding pairs derived from 1K layouts in the CoDeLayout training set, without relying on any external data. Additionally, the 30\% subset alleviates type imbalance, yielding higher Morphing accuracy than MASON (Full data).

\input{tables/moduleablation}

\subsection{Ablation Study}

\subsubsection{Module Ablation}

To assess the contribution of each module, type-balanced subsampling is applied to ensure equal sample counts across types. 
As shown in Table~\ref{tab:ablation}, both multimodal alignment (MA) and structural perception (SP) consistently improve weighted and average accuracy. Direct Finetune (DF) achieves 80.21\% weighted and 78.36\% average accuracy. Adding MA increases performance to 82.03\% and 80.32\%, indicating improved element-level alignment. Incorporating SP yields 83.85\% and 82.51\%, reflecting the benefit of modeling inter-element spatial structure. Combining both modules results in the full MASON model, achieving 84.11\% weighted and 83.40\% average accuracy, demonstrating complementary improvements in addressing semantic drift and structural ambiguity. 
Furthermore, the effects of MA and SP vary across composition types. For heavily occluded compositions (Overlaying/Clipping), noisy grounding may conflict with SP's structural cues, limiting the benefits of MA. In contrast, MA complements SP for Blending. For Morphing, MA alone may increase ambiguity by grounding distorted text as graphics without sufficient structural context, whereas SP helps recover the text-replacement relation.

\input{figs/datascale}

\subsubsection{Data Scale Ablation}
To further demonstrate MASON's data efficiency, we study the effect of training data scale. Direct Finetune and MASON are separately trained on 10\%, 15\%, 30\%, and 100\% of the CoDeLayout training set, with the smaller subsets constructed via type-balanced subsampling. 
As shown in Fig.~\ref{fig:ablation_trends}, performance improves for both methods with increasing training data, while MASON consistently outperforms Direct Finetune across all scales. Under the 10\% setting, Direct Finetune suffers a notable drop in accuracy, whereas MASON maintains a clear advantage, indicating stronger data efficiency. Moreover, while Direct Finetune shows diminishing returns beyond 30\% of the training data, MASON continues to benefit from additional supervision, demonstrating better scalability.

\input{tables/GroundingVLMablation}
\subsubsection{Grounding Model Ablation}
To determine whether the gains from MA stem from the alignment paradigm rather than the strength of a particular grounding model, we conduct an ablation using different grounding VLMs. 
In addition to GPT-4o, we evaluate Qwen3-VL (8B) under two settings:
(1) Direct grounding, where the model generates both element descriptions and positional information; and
(2) Script-based grounding, where positional information is extracted from metadata and the VLM generates only the semantic description. 
As shown in Table~\ref{tab:grounding_vlm_ablation}, script-based grounding achieves performance comparable to GPT-4o. This suggests that the improvements from MA are primarily attributable to explicit metadata–vision alignment rather than dependence on a stronger grounding model.

\subsubsection{Visual Dependency Ablation}
To examine whether the model tends to over-rely on metadata, we evaluate Direct Finetune and MASON with and without visual input. 
As shown in Table~\ref{tab:visual_input_ablation}, removing visual features leads to a substantial performance drop for both methods. This result suggests that compositional layout understanding relies critically on visual perception and cannot be resolved through metadata or layer-aware spatial relationships alone.

\input{figs/casestudy}
\section{Case Study}
To illustrate MASON's advantages over Direct Finetune (DF) in compositional layout understanding, we compare their predictions on representative examples. As shown in Fig.~\ref{fig:case-study}, DF often struggles to localize and associate the correct compositional elements when spatial or semantic cues are subtle, whereas MASON successfully identifies the underlying compositional pairs. 
For example, in the structurally complex Clipping case, DF fails to recognize the implicit relationship between the cutout and the base image, instead selecting a decorative element whose width matches the base image. MASON, by contrast, correctly identifies the cutout element. In the challenging Morphing case, MASON recognizes that the flower image placed over the letter ``O'' creates the morphing effect, whereas DF selects a decorative element aligned with the text and sharing a similar color, but unrelated to the underlying compositional relation.

%% file: tables/main-results.tex
\begin{table}[t]
\centering
\caption{
Performance on CoDeLayout across four compositional categories. Each cell reports Accuracy (GPT-Score/BLEU/ROUGE). Accuracy ($\uparrow$) evaluates compositional element identification, while GPT-Score/BLEU/ROUGE ($\uparrow$) measure explanation quality. \textbf{Bold} and \underline{underline} indicate the best and second-best results. Weighted Accuracy is computed using test sample proportions, and Average Accuracy is the unweighted mean across categories. 
The results highlight challenges in compositional layout understanding for existing VLMs and MASON's effectiveness and data efficiency.
}

\scriptsize

\resizebox{\textwidth}{!}{%
\begin{tabular}{l|cccc|cc}
\toprule
\textbf{Model} &
\textbf{Overlaying} & \textbf{Clipping} & \textbf{Blending} & \textbf{Morphing} & \textbf{Weighted Acc.} & \textbf{Average Acc.} \\
\midrule
\multicolumn{7}{l}{\textbf{Heuristic Baselines}} \\
\midrule
Max-Overlap & \makecell{61.74 \\ \tiny \textcolor{gray}{(N/A)}} & \makecell{69.70 \\ \tiny \textcolor{gray}{(N/A)}} & \makecell{23.08 \\ \tiny \textcolor{gray}{(N/A)}} & \makecell{36.17 \\ \tiny \textcolor{gray}{(N/A)}} & 44.27 & 47.67 \\
Nearest-Neighbor & \makecell{58.26 \\ \tiny \textcolor{gray}{(N/A)}} & \makecell{62.12 \\ \tiny \textcolor{gray}{(N/A)}} & \makecell{23.72 \\ \tiny \textcolor{gray}{(N/A)}} & \makecell{38.30 \\ \tiny \textcolor{gray}{(N/A)}} & 42.44 & 45.60 \\
\midrule
\multicolumn{7}{l}{\textbf{Open-Source VLMs}} \\
\midrule
Qwen2.5-VL & \makecell{66.09 \\ \tiny \textcolor{gray}{(3.51/3.65/23.94)}} & \makecell{37.88 \\ \tiny \textcolor{gray}{(2.85/1.16/18.94)}} & \makecell{53.85 \\ \tiny \textcolor{gray}{(3.37/2.48/22.28)}} & \makecell{36.17 \\ \tiny \textcolor{gray}{(3.22/2.16/21.42)}} & 52.60 & 48.49 \\
Qwen3-VL & \makecell{86.09 \\ \tiny \textcolor{gray}{(4.52/7.63/32.77)}} & \makecell{71.21 \\ \tiny \textcolor{gray}{(3.85/3.81/26.12)}} & \makecell{79.49 \\ \tiny \textcolor{gray}{(4.11/4.59/28.55)}} & \makecell{55.32 \\ \tiny \textcolor{gray}{(3.49/2.22/23.04)}} & 77.08 & 73.02 \\
InternVL-3.5 & \makecell{84.35 \\ \tiny \textcolor{gray}{(4.02/5.66/28.84)}} & \makecell{54.55 \\ \tiny \textcolor{gray}{(2.93/1.29/19.49)}} & \makecell{70.51 \\ \tiny \textcolor{gray}{(3.52/2.74/23.38)}} & \makecell{42.55 \\ \tiny \textcolor{gray}{(3.11/1.70/20.48)}} & 68.48 & 62.99 \\
LLaVA-OneVision & \makecell{56.52 \\ \tiny \textcolor{gray}{(3.41/2.90/23.21)}} & \makecell{21.21 \\ \tiny \textcolor{gray}{(2.72/0.71/17.59)}} & \makecell{60.90 \\ \tiny \textcolor{gray}{(3.24/1.74/21.36)}} & \makecell{29.79 \\ \tiny \textcolor{gray}{(2.89/0.97/18.82)}} & 48.95 & 42.10 \\
\midrule
\multicolumn{7}{l}{\textbf{Closed-Source VLMs}} \\
\midrule
GPT-4o & \makecell{83.48 \\ \tiny \textcolor{gray}{(4.91/10.67/36.34)}} & \makecell{36.36 \\ \tiny \textcolor{gray}{(4.02/3.22/28.05)}} & \makecell{63.46 \\ \tiny \textcolor{gray}{(4.66/6.22/34.00)}} & \makecell{65.96 \\ \tiny \textcolor{gray}{(4.22/5.88/29.89)}} & 65.10 & 62.31 \\
GPT-5 & \makecell{89.57 \\ \tiny \textcolor{gray}{(3.62/4.55/23.95)}} & \makecell{62.12 \\ \tiny \textcolor{gray}{(2.99/1.40/19.18)}} & \makecell{79.49 \\ \tiny \textcolor{gray}{(3.25/2.20/21.25)}} & \makecell{56.92 \\ \tiny \textcolor{gray}{(3.51/1.98/23.97)}} & 76.56 & 71.62 \\
GPT-o3 & \makecell{\underline{93.91} \\ \tiny \textcolor{gray}{(3.35/3.26/22.08)}} & \makecell{63.64 \\ \tiny \textcolor{gray}{(2.81/1.25/18.46)}} & \makecell{79.49 \\ \tiny \textcolor{gray}{(3.12/2.25/19.93)}} & \makecell{68.09 \\ \tiny \textcolor{gray}{(3.46/1.87/23.73)}} & 79.68 & 76.28 \\
Gemini-2.5-flash & \makecell{81.74 \\ \tiny \textcolor{gray}{(4.36/6.88/31.18)}} & \makecell{53.03 \\ \tiny \textcolor{gray}{(3.54/2.96/23.87)}} & \makecell{72.44 \\ \tiny \textcolor{gray}{(3.78/3.61/25.14)}} & \makecell{55.32 \\ \tiny \textcolor{gray}{(3.81/2.85/25.66)}} & 69.79 & 65.63 \\
Gemini-2.5-pro & \makecell{90.43 \\ \tiny \textcolor{gray}{(4.02/4.92/28.48)}} & \makecell{45.45 \\ \tiny \textcolor{gray}{(3.77/2.99/25.79)}} & \makecell{75.00 \\ \tiny \textcolor{gray}{(3.89/3.53/26.33)}} & \makecell{59.57 \\ \tiny \textcolor{gray}{(3.79/3.15/25.70)}} & 72.65 & 67.61 \\
\midrule
\multicolumn{7}{l}{\textbf{Ours}} \\
\midrule
Direct Finetune (Full data) & \makecell{93.04 \\ \tiny \textcolor{gray}{(6.42/30.87/51.63)}} & \makecell{83.33 \\ \tiny \textcolor{gray}{(4.73/10.89/35.08)}} & \makecell{\underline{94.87} \\ \tiny \textcolor{gray}{(5.14/11.69/39.53)}} & \makecell{65.96 \\ \tiny \textcolor{gray}{(4.96/11.23/37.00)}} & 88.80 & 84.30 \\
MASON (30\% data) & \makecell{92.17 \\ \tiny \textcolor{gray}{(6.29/29.92/50.22)}} & \makecell{\underline{86.36} \\ \tiny \textcolor{gray}{(4.81/10.69/35.68)}} & \makecell{93.59 \\ \tiny \textcolor{gray}{(4.86/9.80/36.34)}} & \makecell{\textbf{72.34} \\ \tiny \textcolor{gray}{(4.69/9.95/34.44)}} & \underline{89.32} & \underline{86.12} \\
MASON (Full data) & \makecell{\textbf{95.65} \\ \tiny \textcolor{gray}{(6.62/31.87/52.99)}} & \makecell{\textbf{90.91} \\ \tiny \textcolor{gray}{(4.93/11.74/36.27)}} & \makecell{\textbf{95.51} \\ \tiny \textcolor{gray}{(5.22/11.84/39.66)}} & \makecell{\underline{70.21} \\ \tiny \textcolor{gray}{(5.13/11.60/38.77)}} & \textbf{91.66} & \textbf{88.07} \\
\bottomrule
\end{tabular}%
}
\label{tab:main_results}
\end{table}

%% file: tables/moduleablation.tex
\begin{table}[t]
\centering
\caption{
Module ablation on CoDeLayout under a type-balanced setting. 
DF denotes Direct Finetune, MA denotes Multimodal Alignment, and SP denotes Structural Perception. Results report accuracy ($\uparrow$) for each compositional category, along with Weighted and Average accuracy. Weighted accuracy accounts for category proportions, while Average accuracy is the unweighted mean. \textbf{Bold} indicates the best result in each column. Results show that MA and SP provide complementary improvements, with their combination achieving the strongest overall performance.
}

\scriptsize

\begin{tabular*}{\linewidth}{@{\extracolsep{\fill}}lcccccc@{}}
\toprule
\textbf{Model} & 
\textbf{Overlaying} & 
\textbf{Clipping} & 
\textbf{Blending} & 
\textbf{Morphing} & 
\textbf{Weighted Acc.} & 
\textbf{Average Acc.} \\
\midrule
DF & 86.09 & 74.24 & 80.77 & 72.34 & 80.21 & 78.36 \\
DF + MA & 87.83 & 81.82 & 81.41 & 70.21 & 82.03 & 80.32 \\
DF + SP & \textbf{89.57} & \textbf{83.33} & 82.69 & 74.47 & 83.85 & 82.51 \\
MASON & 86.96 & 81.82 & \textbf{83.97} & \textbf{80.85} & \textbf{84.11} & \textbf{83.40} \\
\bottomrule
\end{tabular*}

\label{tab:ablation}
\end{table}

%% file: figs/datascale.tex
\begin{figure}[t]
    \centering
    \begin{minipage}[c]{0.42\linewidth}%
        \centering
        \includegraphics[width=\linewidth]{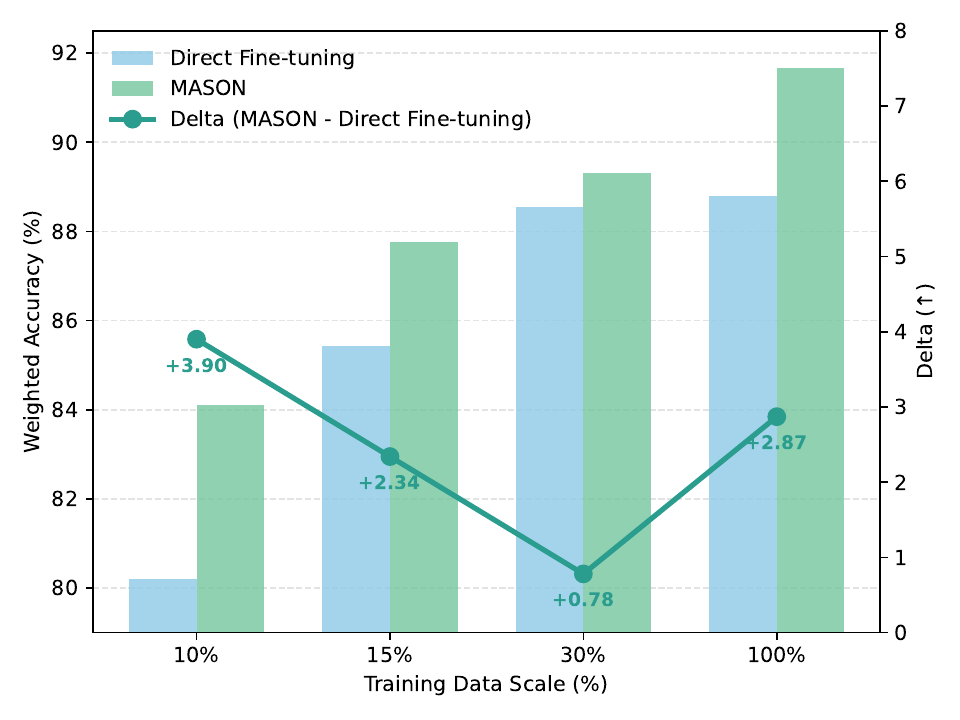}
    \end{minipage}%
    \hfill%
    \begin{minipage}[c]{0.56\linewidth}%
        \caption{
        \textbf{Data Scale Ablation.} 
        Performance of Direct Finetune and MASON trained on 10\%, 15\%, 30\%, and 100\% of the CoDeLayout training set. The x-axis denotes training data scale, the left y-axis shows test set weighted accuracy, and the right y-axis shows accuracy gain of MASON over Direct Finetune. MASON consistently outperforms across all scales, with a clear advantage under low-resource settings.
        }
        \label{fig:ablation_trends}
    \end{minipage}
\end{figure}

%% file: tables/GroundingVLMablation.tex
\begin{table}[t]
\centering
\scriptsize

\begin{minipage}[t]{0.5\linewidth}
\centering
\caption{
\textbf{Grounding Model Ablation.} Comparable performance across grounding models suggests that the gains arise from the alignment paradigm rather than model strength.
}
\label{tab:grounding_vlm_ablation}
\setlength{\tabcolsep}{2pt}
\scriptsize
\begin{tabular}{@{}lcc@{}}
\toprule
Grounding VLM & Weighted Acc. & Average Acc. \\
\midrule
GPT-4o            & 91.6\% & 88.1\% \\
Qwen3-VL (direct) & 90.7\% & 87.4\% \\
Qwen3-VL (script) & 91.1\% & 88.1\% \\
\bottomrule
\end{tabular}
\end{minipage}
\hfill
\begin{minipage}[t]{0.48\linewidth}
\centering
\caption{
\textbf{Visual Dependency Ablation.} Removing visual features leads to a substantial performance drop, highlighting the importance of visual perception for compositional layout understanding.
}
\label{tab:visual_input_ablation}
\setlength{\tabcolsep}{2pt}
\makebox[\linewidth][c]{%
\scriptsize
\begin{tabular}{@{}lcccc@{}}
\toprule
 & \multicolumn{2}{c}{Weighted Acc.} & \multicolumn{2}{c}{Average Acc.} \\
\cmidrule(lr){2-3} \cmidrule(lr){4-5}
Model & Visual & No Visual & Visual & No Visual \\
\midrule
DF    & 88.8\% & 60.9\% & 84.3\% & 58.1\% \\
MASON & 91.6\% & 65.1\% & 88.0\% & 62.8\% \\
\bottomrule
\end{tabular}}
\end{minipage}

\end{table}

%% file: figs/casestudy.tex
\begin{figure*}[t]
    \centering

    \begin{subfigure}[t]{0.27\textwidth}
        \centering
        \includegraphics[width=\linewidth]{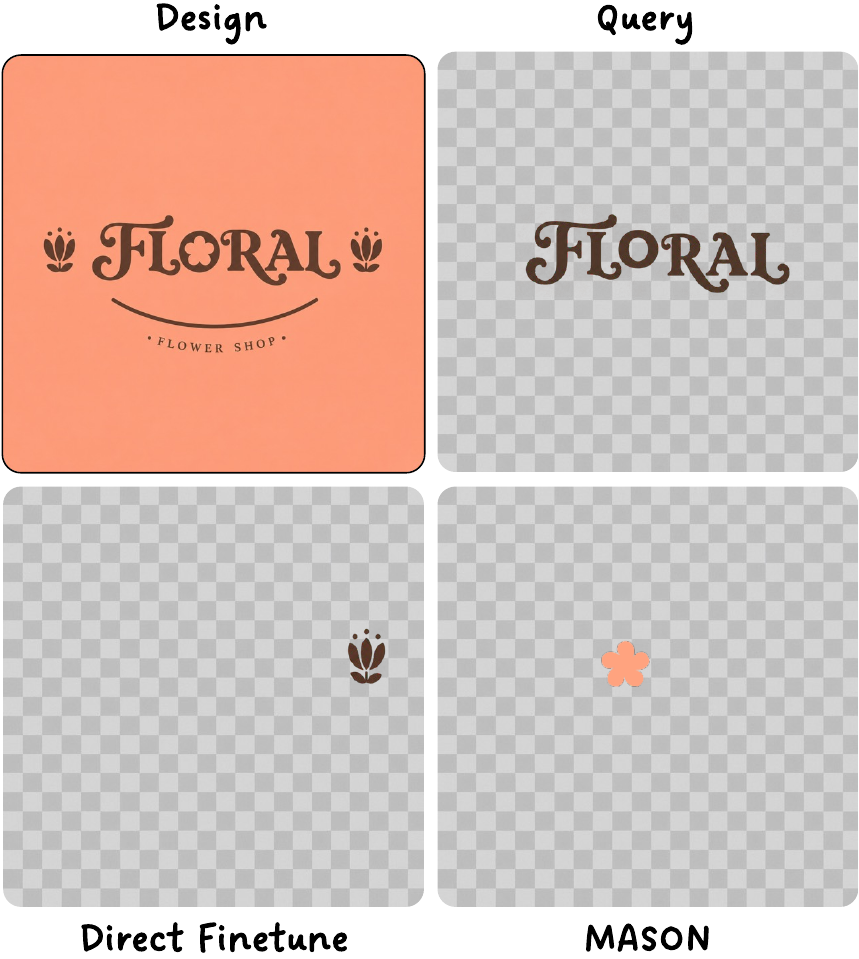}
        \caption{Morphing}
        \label{fig:case-morphing}
    \end{subfigure}
    \hfill
    \begin{subfigure}[t]{0.2115\textwidth}
        \centering
        \includegraphics[width=\linewidth]{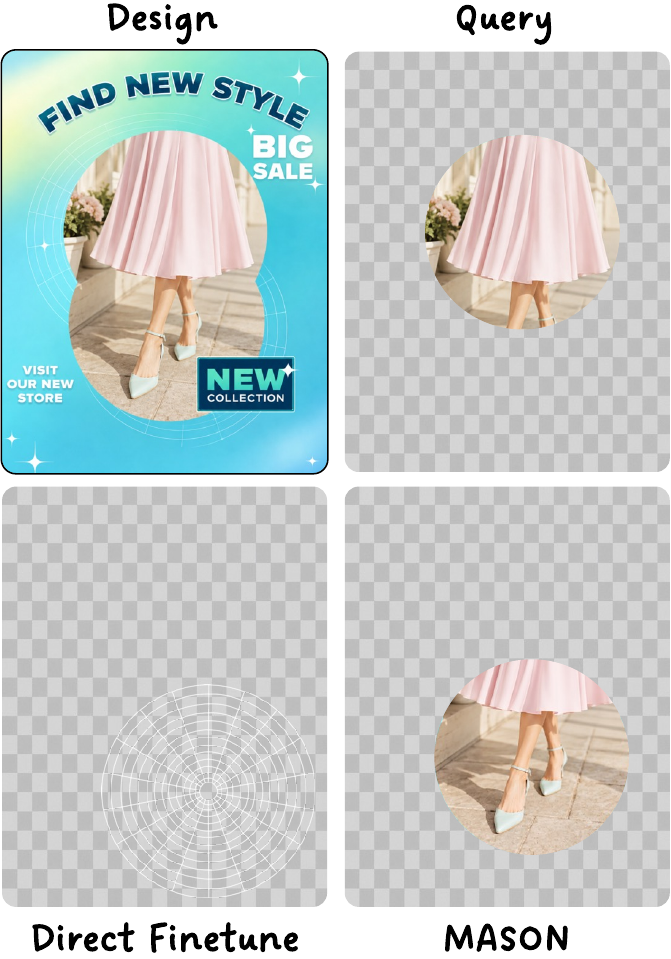}
        \caption{Blending}
        \label{fig:case-blending}
    \end{subfigure}
    \hfill
    \begin{subfigure}[t]{0.1745\textwidth}
        \centering
        \includegraphics[width=\linewidth]{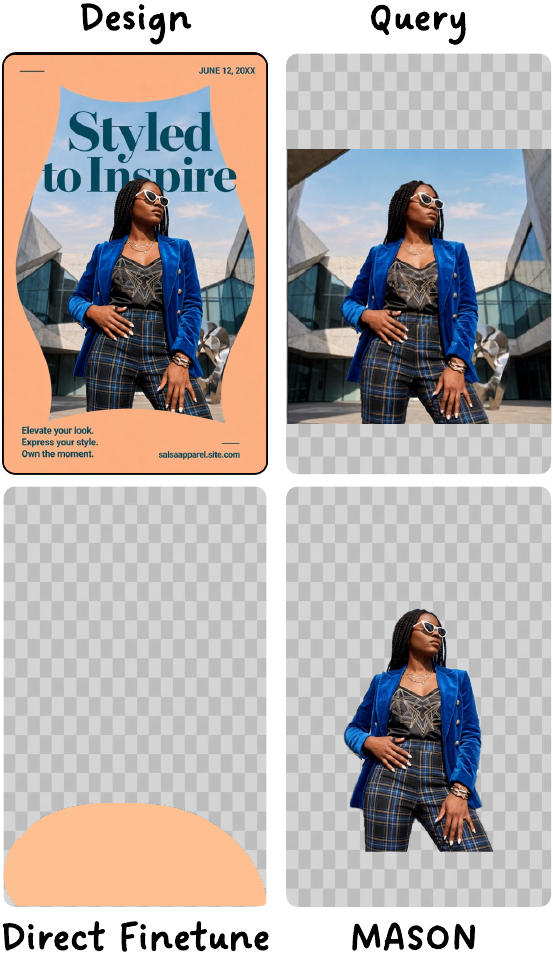}
        \caption{Clipping}
        \label{fig:case-clipping}
    \end{subfigure}
    \hfill
    \begin{subfigure}[t]{0.27\textwidth}
        \centering
        \includegraphics[width=\linewidth]{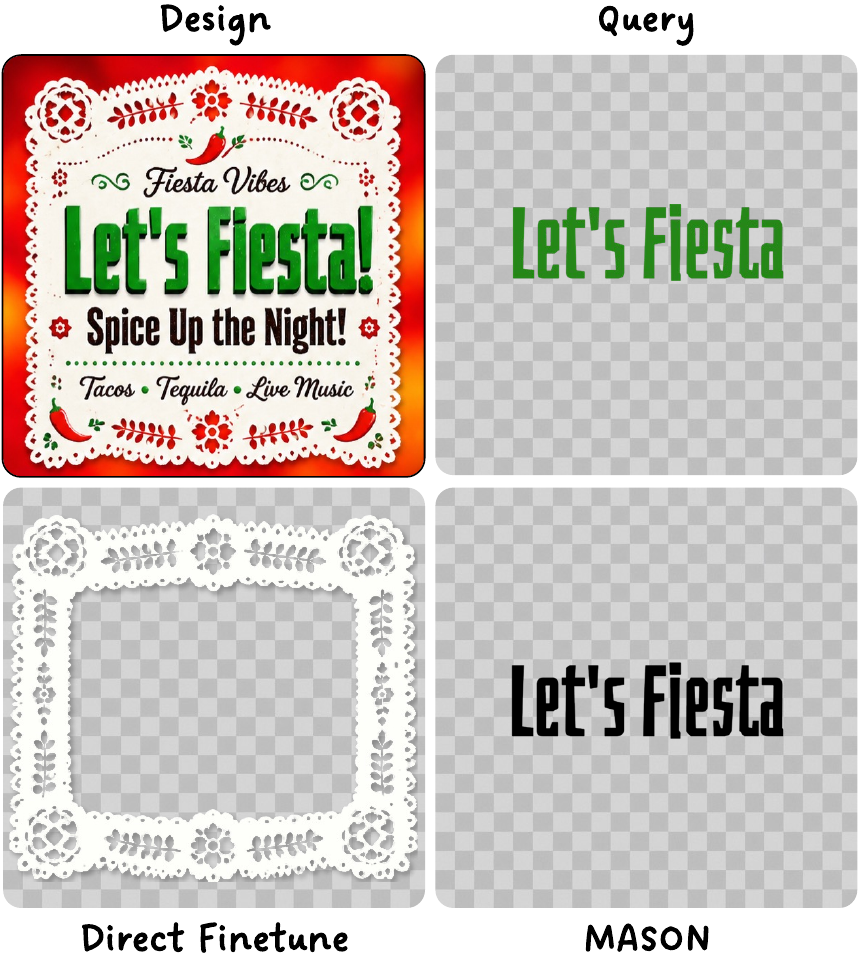}
        \caption{Overlaying}
        \label{fig:case-overlaying}
    \end{subfigure}

    \caption{Case studies across four compositional categories. Each example shows the full design, the query element, and the predictions of Direct Finetune and MASON.}
    \label{fig:case-study}
\end{figure*}

%% file: sec/5_conclusion.tex
\section{Conclusion}
This paper introduces compositional design layout understanding, a new task that challenges VLMs to interpret visually entangled elements and model layer-aware inter-element relationships within hierarchical layouts, and presents CoDeLayout, the first dataset dedicated to this setting. Through empirical analysis, we identify two core challenges: \textit{semantic drift}, where VLMs fail to align textual metadata with visual content, and \textit{structural ambiguity}, where VLMs struggle to capture hierarchical and spatial relationships among elements. To address these challenges, we propose MASON, a paradigm that integrates multimodal alignment and structural perception into VLM post-training. By introducing grounding-based alignment supervision and incorporating layer-aware spatial relationships, MASON enhances both element-level interpretation and inter-element structural modeling, thereby improving compositional layout understanding. Extensive experiments show that MASON consistently outperforms both open- and closed-source VLMs, including reasoning-capable models, while demonstrating strong data efficiency under limited supervision.

Future work includes developing layer-aware visual encoders tailored to multilayer compositional layouts and exploring stronger cross-modal alignment objectives to further enhance element-level grounding and relational reasoning.